\pdfoutput=1

\documentclass[11pt]{article}

\PassOptionsToPackage{dvipsnames}{xcolor}
\usepackage[preprint]{acl}

\usepackage{times}
\usepackage{latexsym}

\usepackage[T1]{fontenc}

\usepackage[utf8]{inputenc}

\usepackage{microtype}

\usepackage{inconsolata}

\usepackage{graphicx}
\usepackage{subcaption}

\usepackage{xspace}
\usepackage{ntheorem}
\usepackage{amssymb}
\usepackage{amsmath}
\usepackage{xcolor}
\usepackage{cleveref}
\usepackage{booktabs}
\usepackage{array}

\newtheorem{hypoth}{Hypothesis}
\newtheorem{definition}{Definition}

\newcommand{\lang}[0]{\ensuremath{\ell}\xspace}

\newcommand{\Word}{\textcolor{WordsColor}{\ensuremath{\mathrm{W}}}\xspace}
\newcommand{\word}{\textcolor{WordsColor}{\ensuremath{\mathrm{w}}}\xspace}
\newcommand{\vocab}{\textcolor{WordsColor}{\ensuremath{\mathcal{W}}}\xspace}

\newcommand{\Words}{\textcolor{WordsColor}{\ensuremath{\mathbf{W}}}\xspace}
\newcommand{\words}{\textcolor{WordsColor}{\ensuremath{\mathbf{w}}}\xspace}
\newcommand{\Wordsleft}{\textcolor{WordsColor}{\ensuremath{\Words_{\leftarrow}}}\xspace}
\newcommand{\wordsleft}{\textcolor{WordsColor}{\ensuremath{\words_{\leftarrow}}}\xspace}
\newcommand{\Wordsbi}{\textcolor{WordsColor}{\ensuremath{\Words_{\leftrightarrow}}}\xspace}
\newcommand{\wordsbi}{\textcolor{WordsColor}{\ensuremath{\words_{\leftrightarrow}}}\xspace}

\newcommand{\entropy}{\ensuremath{\mathrm{H}}\xspace}
\newcommand{\mi}{{\ensuremath{\mathrm{MI}}}\xspace}

\newcommand{\Pitch}{\textcolor{PitchColor}{\ensuremath{\boldsymbol{\mathrm{P}}}}\xspace}
\newcommand{\pitch}{\textcolor{PitchColor}{\ensuremath{\boldsymbol{\mathrm{p}}}}\xspace}

\newcommand{\params}{\textcolor{AccentColor}{\ensuremath{\theta}}\xspace}
\newcommand{\pdist}{{\ensuremath{\mathcal{Z}}}\xspace}
\newcommand{\lm}{{\ensuremath{\text{LM}_{\params}}}\xspace}
\newcommand{\normal}{{\ensuremath{\mathcal{N}}}\xspace}

\newcommand{\kdeall}{{\ensuremath{\text{KDE-}\Word\textsc{(all)}}}\xspace}
\newcommand{\kdesplit}{{\ensuremath{\text{KDE-}\Word\textsc{(split)}}}\xspace}
\newcommand{\mdnword}{{\ensuremath{\text{MDN-}\Word}}\xspace}
\newcommand{\mdnpast}{{\ensuremath{\text{MDN-}\Wordsleft}}\xspace}
\newcommand{\mdnbi}{{\ensuremath{\text{MDN-}\Wordsbi}}\xspace}

\definecolor{PitchColor}{RGB}{27,153,139}
\definecolor{WordsColor}{RGB}{33,92,175} 
\definecolor{AccentColor}{RGB}{244,96,54}

\newcommand{\defn}[1]{\textbf{#1}}

\title{Using Information Theory to Characterize Prosodic Typology: The Case of Tone, Pitch-Accent and Stress-Accent}

\author{Ethan Gotlieb Wilcox \\
  Georgetown University \\
  \href{mailto:ethan.wilcox@georgetown.edu}{\texttt{ethan.wilcox@georgetown.edu}} \\\And
  Cui Ding \\
  University of Zürich \\
  \href{mailto:cui.ding@uzh.ch}{\texttt{cui.ding@uzh.ch}} \\\And
  Giovanni Acampa \\
  ETH Zürich \\
  \href{mailto:giovanni.acampa@inf.ethz.ch}{\texttt{giovanni.acampa@inf.ethz.ch}} \\\AND 
  Tiago Pimentel \\
  ETH Zürich \\
  \href{mailto:tiago.pimentel@inf.ethz.ch}{\texttt{tiago.pimentel@inf.ethz.ch}}\\\And
  Alex Warstadt \\
  UC San Diego \\
  \href{mailto:awarstadt@ucsd.edu}{\texttt{awarstadt@ucsd.edu}}\\\And
  Tamar I. Regev \\
  MIT \\
  \href{mailto:tamarr@mit.edu}{\texttt{tamarr@mit.edu}}}

\begin{document}

\maketitle

\begin{abstract}

This paper argues that the relationship between lexical identity and prosody---one well-studied parameter of linguistic variation---can be characterized using information theory.
We predict that languages that use prosody to make lexical distinctions should exhibit a higher mutual information between word identity and prosody, compared to languages that do not.
We test this hypothesis in the domain of pitch, which is used to make lexical distinctions in tonal languages, like Cantonese. 
We use a dataset of speakers reading sentences aloud in ten languages across five language families to estimate the mutual information between the text and their pitch curves.
We find that, across languages, pitch curves display similar amounts of entropy.
However, these curves are easier to predict given their associated text in tonal languages, compared to pitch- and stress-accent languages; the mutual information is thus higher in these languages, supporting our hypothesis.
Our results support perspectives that view linguistic typology as gradient, rather than categorical.

\vspace{.2em}
\hspace{1.25em}\includegraphics[width=1.25em,height=1.25em]{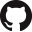}{\hspace{.75em}\parbox{\dimexpr\linewidth-2\fboxsep-2\fboxrule}{\url{https://github.com/picol-georgetown/Prosody_Typology}}}

\end{abstract}

\section{Introduction}

One central tension in linguistics is between linguistic \emph{universality} and \emph{diversity}. The world contains some 7,000 languages \citep{ethnologue2023languages}, each with its unique and idiosyncratic lexicon, phonological inventory, and grammar. At the same time, linguistic properties are shared between sets of related languages \citep{croft2002typology}, and some features appear, or covary, across languages, giving rise to the hypothesis that human language is governed by a set of universal principles \citep{greenberg2005universals}. Major advances in the study of language have been made through the introduction of frameworks that can describe both the typological variation observed between languages as well as the universal consistencies observed across languages. Examples of such frameworks are the Principles and Parameters approach for syntactic structure \citep{chomsky1993lectures, culicover1997principles} and Optimality Theory for phonological systems \citep{prince2004optimality}.

One promising candidate for this type of framework is information theory \citep{shannon1948mathematical}. Studies have argued that information-theoretic approaches can explain universal principles in languages, including the distribution of word lengths \citep{zipf1949human, piantadosi2011word, pimentel-etal-2023-revisiting}, the organization of semantic systems \citep{kemp2018semantic, zaslavsky2018efficient, zaslavsky2021let}, word orders \citep{dyer-etal-2021-predicting} as well as language processing phenomena \citep{futrell2020lossy, wilcox2023testing}. However, information-based approaches are less widely used to describe typological variation (although cf.,  \citealp{futrell2020lossy, pimentel-etal-2020-phonotactic,socolof2022measuring, steuer-etal-2023-information}). In this paper, we take one well-studied crosslinguistic parameter---whether or not a language has lexical tone---and argue that it can be characterized information-theoretically, as the amount of mutual information between a lexical item (i.e., a word) and the pitch curve associated with that word. Our goal is to demonstrate how an information-based approach can be used to characterize crosslinguistic variation, as well as to showcase how NLP methods can be used to formally quantify properties that are debated in the formal linguistics literature, e.g., whether, or to what extent, a given language or dialect is tonal \citep{hyman2006word}.

The domain we are interested in is prosody---the melody of speech. A word’s prosody is transmitted through several unique features, including its duration, energy (perceived as loudness), and fundamental frequency (perceived as pitch). Pitch, specifically, is the main focus of our study. Crucially, the role that pitch plays varies across languages, with phonologists traditionally placing languages in three broad categories: In \defn{tonal languages} such as Vietnamese, Mandarin, and Yoruba, all or most syllables carry one of several discrete pitch contours which differentiate between lexical items; in \defn{stress-accent} languages such as English and Italian pitch does not differentiate between lexical identity at all, playing other roles like providing cues for stress placement, or indicating whether or not a sentence is a question. In an intermediate set of languages, called \defn{pitch-accent} languages, such as Swedish or Japanese, pitch contours are lexically contrastive, but they are not present on every word.

The reason why we focus on pitch and its relationship to tone is that in the phonology literature, this issue has been at the forefront of debates about how one ought to make typological distinctions. Using tonal systems as an example, some have argued that the job of typology is to identify language ``types’’ \citep[i.e., the three in the above paragraph;][]{hagege1992morphological}, while others have argued that typology should be viewed as laying out a typological ``continuum’’ over several prosodic properties \citep{hyman2006word}, and have questioned whether stress-accent languages are a single, naturally-occurring linguistic category. As an example, Western Basque, Tokyo Japanese, and Luganda (a Bantu language) make \emph{some} lexical distinctions based on pitch. However, the number of such words varies between the languages, and their ``tonal'' systems interact idiosyncratically with other aspects of the language's phonology \citep{hyman2006word}. Is it fair to say that these languages belong to a single type? And if they fall on a continuum, then what metric should one use as its basis?

Information-theoretic approaches offer a new way of exploring such continua, and can offer new evidence for or against typological clusters. Our approach offers theoretically motivated quantities, which are estimated from raw audio data, and therefore inherently capture several of the prosodic properties that have been hypothesized to make up tonal continua. Our contribution is in line with several recent studies that have recast other aspects of typological variation in information-theoretic terms, e.g., for morphological fusion \citep{socolof2022measuring} and vowel harmony \citep{steuer-etal-2023-information}. Specifically, we hypothesize that because tonal languages use pitch to distinguish lexical identity, given a lexical item's identity, it should be easier to predict a word's pitch curve in tonal compared to non-tonal languages. Information-theoretically, this means there should be more mutual information between lexical identity and pitch in tonal languages, such as Cantonese, than in non-tonal languages, such as English.

To test this hypothesis, we use a pipeline \citep{wolf-etal-2023-quantifying} originally developed in English to measure the mutual information (\mi) between prosody and written text; where text is used as a proxy for lexical identity. 
We make several technical contributions to this pipeline, enabling it to produce more accurate \mi estimates across languages. We measure mutual information for ten typologically distinct languages: English, French, Italian, German, Swedish (Indo-European), Mandarin, Cantonese (Sino-Tibetan), Japanese (Japonic), Thai (Kra–Dai), and Vietnamese (Austroasiatic). 
These languages are traditionally classified as either stress-accent, pitch-accent, or tonal. We find that, across languages, pitch curves display similar amounts of entropy, suggesting that the information conveyed by the pitch channel is conserved cross-linguistically. 
However, these curves are easier to predict given their associated text in the tonal languages, compared to pitch- and stress-accent languages, and thus the \mi is higher in these languages, supporting our hypothesis.
Interestingly, the mutual information does not follow a multimodal distribution, which would classify languages into clearly distinct categories. Rather, they show a continuum of values, in line with perspectives favoring a gradient, rather than a categorical approach to prosodic typology \citep{hyman2006word} and linguistic typology more broadly \citep{pimentel-etal-2020-phonotactic,levshina2023gradient, baylor2024multilingual}.

\section{Prosodic Typology}

In this section, we provide a formal framework for describing linguistic typologies based on prosodic features. We start by outlining our notation: We assume that each natural language consists of lexical items, \word, drawn from a lexicon \vocab. We use \Word to denote a lexical-item-valued random variable. By ``lexical item'' we mean the sense of dictionary definitions---each value of \Word is associated with a unique lexical item, rather than with a particular orthographic representation. However, as we do not have direct access to lexical identities in a large corpus, we will relax this in our experiments and work instead with orthographic words, which we use as a proxy for lexical identities. In addition, we define \pitch, as a real-valued vector that represents 
some prosodic feature for a given word. Although in our subsequent experiments, \pitch refers only to the pitch curve, for now we will use \pitch for prosody as a whole, including other features, such as average acoustic energy or duration. We denote a prosody-valued random variable as \Pitch.

What does it mean for a language to have contrastive tone, stress, or length?
In linguistics textbooks, this is often defined through minimal pairs, by showing that there are systematic correspondences between lexical identity and the prosodic feature of interest. For example, \citet{yip2002tone} illustrates the notion of a tonal language by giving an example of the syllable \emph{[yau]} in Cantonese. If pronounced with a high-rising tone, this syllable means \emph{paint}; however, if pronounced with a low-level tone, it means \emph{again}. 
Based on such examples, we propose the following definition:

\begin{definition} \label{def:prosody-lang}
    A language \lang is typologically a \pitch-language if, in \lang, prosodic feature \pitch provides information about lexical identity.
\end{definition}

That is, if a language is a \pitch-language, then knowing the prosodic value, \pitch, of a particular lexical item, \word, should make that word easier to predict. As an example, in Cantonese, if we know a word has a high-rising tone, then it will be easier to predict that word's meaning, compared to a situation where we don't know the pitch at all.\footnote{We acknowledge that ``providing information about'' lexical identity is a less stringent requirement than, say \emph{determining} lexical identity. We adopt this definition, in part, because it is more conducive to measuring experimentally.}

Based on this definition, we propose that one natural way to describe prosodic typologies is through the lens of information theory.
Under information theory, if a variable (e.g., $\Pitch$) makes another variable (e.g., $\Word$) easier to predict, we say that it contains information about it.
We can thus say that a $\pitch$-language should be one where pitch conveys information about lexical identity, written as:
\begin{align} \label{eq:mi-inequality}
    \mi(\Pitch;\Word) > 0
\end{align}

That is, the \defn{mutual information (\mi)} between $\pitch$ and $\word$ is greater than zero. Conversely, in non \pitch-languages, where $\pitch$ does not determine lexical identity, the mutual information will be roughly equivalent to zero, i.e., $\mi(\Pitch;\Word) \approx 0$. Note that because mutual information is symmetric, in  $\pitch$-languages, we also predict that lexical identity reduces uncertainty about prosodic features, which is what we empirically test in the following sections.

\subsection{Predictions: Tone, Stress and Pitch-accent} \label{sec:predictions}

The prediction outlined in \cref{eq:mi-inequality} is limited in several ways.
First, it predicts that the \mi in non-tone languages should be no different from zero. However, as noted above, even in stress-accent languages, pitch can carry indirect information about lexical identity.
Second, the prediction only makes a binary classification: \mi should be positive in \pitch-languages, and equal to zero in non \pitch-languages. However, in real life, we expect that things are more complicated. Rather than a single distinction, one might expect to find more nuanced differences between languages. This should be the case especially when it comes to pitch---the focus of our study---as existing typologies already separate languages into (at least) three categories based on the relationship between pitch and lexical identity. We therefore outline three more concrete hypotheses concerning the mutual information, \mi, of a language's lexical identity (\Word) and pitch (\Pitch):\looseness=-1

\begin{hypoth} \label{hyp:ordering}
    Typological Ordering Hypothesis: Languages will display the following ordering of average \mi within linguistic typological groups: tonal languages >> pitch-accent languages >> stress-accent languages.
\end{hypoth}

\noindent In addition, we formulate two competing hypotheses that correspond to different approaches toward linguistic typology:

\begin{hypoth}
    Categorical Prediction: Languages will display a categorical distinction in MI, divided into modes corresponding to typological groups.
\end{hypoth}

\begin{hypoth}
    Gradient Prediction: Languages will display a gradient in \mi on a continuum. Differences between languages within a typological group can be as large as differences across groups.
\end{hypoth}

\noindent To explore these hypotheses, we improve an existing pipeline for estimating \mi, the details of which we will turn to in \cref{sec:methods}.

\subsection{A Type- or Token-level Prediction?}

It is important to note the nature of the information we treat here.
In particular, we could define the \mi above in two ways: at the type or token level.
These would quantify categorically different linguistic properties.
A type-level $\mi(\Pitch;\Word)$ measures how predictable a novel word's pitch is given its lexical identity; it would thus quantify if $\pitch$-values are systematically assigned to words based on their meaning or orthography. 
As lexicons' form--meaning assignments are largely arbitrary  \citep[a property known as the arbitrariness of the sign;][]{saussure1916course,dautriche2017wordform,pimentel-etal-2019-meaning}, we would expect such type-level \mi to be small in both $\pitch$- and non-$\pitch$-languages.
A token-level $\mi(\Pitch; \Word)$, on the other hand, quantifies how well $\pitch$ disambiguates \emph{known} words in a language, and should thus have significantly different values in $\pitch$- and non-$\pitch$-languages.
We thus focus on this \mi's token-level definition here.

\section{Methods}\label{sec:methods}

The prediction in \cref{eq:mi-inequality} is about \emph{lexical items}, however, we do not have direct access to these in the multilingual corpora we use for this study. Rather, we have access to textual representations, i.e., orthographic words, which often correspond to lexical items. In the rest of this paper, therefore, we take \Word to be a random variable corresponding to either a piece of \emph{text} or an orthographic word. Furthermore, as we are specifically interested in pitch, from here on $\Pitch$ is a random variable that represents the parameterization of a pitch curve, specifically, as opposed to just a general prosodic feature. We discuss how we represent \Pitch at greater length in \Cref{sec:pitch-repr}.

\subsection{Estimating Mutual Information}

We estimate the mutual information between prosody and text, following the proposal from \citet{wolf-etal-2023-quantifying}.
\citeauthor{wolf-etal-2023-quantifying} estimate this quantity by first decomposing \mi as the difference between two entropies, and separately estimating each term
\begin{subequations}
\begin{align}
    \mi(\Pitch, \Word) & = \entropy(\Pitch) - \entropy(\Pitch \mid \Word) \label{eq:mi-decomp} \\
    & \approx \entropy_{\params}(\Pitch) - \entropy_{\params}(\Pitch \mid \Word) \label{eq:mi-decom-estimation}
\end{align}
\end{subequations}

As represented by \cref{eq:mi-decom-estimation}, we estimate the \mi as the difference between two \emph{cross entropies}, $\entropy_{\params}(\cdot)$. The cross-entropy is defined as the expectation of $- \log p_{\params}(\pitch)$ or $- \log p_{\params}(\pitch \mid \word)$, given the ground-truth distribution $p(\pitch)$ or $p(\pitch \mid \word)$, respectively. Following \citet{wolf-etal-2023-quantifying}, we use \emph{redistributive sampling} \citep{tibshirani1993introduction, beirlant1997nonparametric} to estimate these quantities. Given model $p_{\params}$, we select a set of $N$ held-out test samples from our dataset, and then estimate each quantity as the average negative log probability (i.e., surprisal) of these test items:
\begin{subequations}
\begin{align}
    &\entropy_{\params}(\Pitch) \approx \frac{1}{N} \sum_{n=1}^{N} \log \frac{1}{p_{\params}(\pitch^n)} \label{eq:unconditional_entropy} \\
    &\entropy_{\params}(\Pitch \mid \Word) \approx \frac{1}{N} \sum_{n=1}^{N} \log \frac{1}{p_{\params}(\pitch^n \mid \word^n)}
\end{align}
\end{subequations}
Where $\pitch^n$ and $\word^n$ are the $n^{th}$ text/pitch pair in our test set. In order to make this estimation, we need to learn a probability distribution $p_{\params}(\pitch)$ and $p_{\params}(\pitch \mid \word)$. We do so with the following methods.

\subsubsection{Estimating \texorpdfstring{$p_{\params}(\pitch)$}{Unconditional Model}} Following \citet{wolf-etal-2023-quantifying} we estimate the unconditional distribution with a Gaussian Kernel Density Estimate, KDE \citep{parzen1962estimation, sheather2004density}. Bandwidth is optimized via $10$-fold cross-validation, using the training and validation data, selecting from Scott’s rule, Silverman’s rule, and fixed values \citep{silverman1986density}. We implement this with \texttt{SciPy} \citep{virtanen2020scipy}. After selecting the optimal bandwidth, we fit the KDE on the training data and compute \cref{eq:unconditional_entropy} on the held-out test data.\looseness=-1

\subsubsection{Estimating \texorpdfstring{$p_{\params}(\pitch \mid \word)$}{Conditional Model}} 

\citet{wolf-etal-2023-quantifying} estimate this conditional distribution by using a neural network to \emph{learn} the parameterization, $\mathbf{\phi}$ of a predictive distribution $\pdist_{\mathbf{\phi}}(\cdot)$ that captures the desired conditional probability distribution, $p_{\params}(\pitch \mid \word)$. In their setup, the predictive distribution is always either a Gaussian or Gamma distribution. This, however, leads to a discrepancy between the expressivity of the distribution learned for the conditional and unconditional distributions, $p_{\params}(\pitch \mid \word)$ and $p_{\params}(\pitch)$. 
The KDEs used to model $p_{\params}(\pitch)$ construct \emph{non-parametric} distributions from the bottom-up, summing together many Gaussians and having a number of parameters that grows with $K$, the number of samples in the training dataset; this distribution can thus be increasingly complex given larger training datasets. 
However, the learned conditional distribution $p_{\params}(\pitch \mid \word)$  in \citet{wolf-etal-2023-quantifying}, is fit as a \emph{parametric} distribution $\pdist$ (Gaussian or Gamma), and is thus constrained independently of the training dataset size.
Therefore, the two distributions do \emph{not} allow for an apples-to-apples comparison. In particular, the greater expressivity of the unconditional distribution $p_{\params}(\pitch)$ means that, in practice, \cref{eq:mi-decom-estimation} is likely to underestimate the true mutual information and can even be negative. To fix this problem, we use two different methods for estimating the conditional probability distribution with greater expressivity, which we outline below.

\paragraph{Conditional KDEs:}

For this method, we partition the dataset by orthographic word type and fit a different KDE for each partition. The resulting KDE is conditionalized on a given word insofar as it has seen only examples of that word's prosody during the estimation procedure. We use two different estimation procedures: In the first, \kdeall,
we use the whole dataset for bandwidth selection, training, and entropy estimation. In the second, \kdesplit, we use 70\% of the dataset for bandwidth selection and training and estimate entropy using redistributive sampling on the held-out portion. One issue with this method is that if a word has relatively few samples in our training data, then our fitted conditional KDE estimate will not be very accurate. To alleviate this problem, we select a threshold $\lambda$; for words that occur fewer times than this threshold, we set their probability to be that of the \emph{unconditional} KDE model. This backoff strategy effectively sets the Pointwise Mutual Information (PMI) to be zero for these words in the final \mi calculation.
We conducted several pilot experiments with $\lambda=\{20, 30, 40, 50, 60\}$ and found that the qualitative nature of the results did not change. 
In Section~\ref{sec:results}, we present results for $\lambda=20$.\looseness=-1

\newcommand{\hiddenstates}{\mathbf{h}}
\newcommand{\mlpfunc}{\mathtt{mlp}}

\paragraph{Mixture Density Networks (MDNs):}

For our second method, we employ a \defn{mixture density network} \citep[MDN;][]{bishop1994mixture}. MDNs are very similar to KDE estimators insofar as the final conditional probability is the sum of several Gaussian kernels. However, the means and variances of these Gaussians are \emph{learned} by a neural network, \lm, with parameters \params, given input \word. 
In addition, the network also learns a set of weights $w^k$ that govern the mixture of the individual Gaussians. The conditional distribution is therefore:
\begin{align}
    & p_{\params}(\pitch \mid \word) = \sum_{k=1}^K w^k_{\word;\params} \: \normal(\pitch \mid \boldsymbol{\mu}^k_{\word;\params}, \boldsymbol{\Sigma}^k_{\word;\params})
\end{align}
where $w^k$ is the weight, $\boldsymbol{\mu}^k$ (a vector in $\mathbb{R}^d$, where $d=4$ is the dimension of $\pitch$, see \Cref{sec:pitch-repr}) is the mean and $\boldsymbol{\Sigma}^k$ (a diagonal co-variance matrix in $\mathbb{R}^{d\times d}$, assuming independence between the different dimensions of \pitch) is the variance of the $k^{th}$ Gaussian kernel parameterized by \params given input \word. 
We use $K = 20$ kernels. 
The MDN itself consists of multilayer perceptrons which receive fastText representations \citep{bojanowski-etal-2017-enriching} and output the mixture of Gaussians' parameters.
Writing these representations as $\hiddenstates \in \mathbb{R}^{d_{\mathrm{ft}}}$:%
\begin{subequations}
\begin{align}
    w^k_{\word;\params} = \mlpfunc_{w^k}(\hiddenstates), \\
    \boldsymbol{\mu}^k_{\word;\params} = \mlpfunc_{\boldsymbol{\mu}^{k}}(\hiddenstates), \\
    \boldsymbol{\Sigma}^k_{\word;\params} = \mlpfunc_{\boldsymbol{\Sigma}^{k}}(\hiddenstates)\,
\end{align}
\end{subequations}
where the number of hidden layers and hidden units in these MLPs are hyperparameters.
Details of our hyperparameter search are given in Appendix \ref{app:hyperparameter}.
We refer to this method as \mdnword.

\subsection{Estimating the MI between Prosody and Longer Textual Contexts} \label{sec:mi_pitch_sentence}

Beyond the $\mi(\Pitch; \Word)$ between prosody and a lexical item, we will also analyse two other mutual informations:
the \mi between pitch and a word's autoregressive (i.e., previous) context, $\mi(\Pitch; \Wordsleft)$; and
the \mi between pitch and a word's bidirectional context, $\mi(\Pitch; \Wordsbi)$.
We estimate these values using MDNs identical to those in the previous section, but these MDNs receive as input representations from mGPT \citep{shliazhko-etal-2024-mgpt}, a multilingual autoregressive language model, largely based on the GPT-2 architecture, or from mBERT \citep{devlin-etal-2019-bert}, a multilingual version of BERT. 
These MDNs give us estimates of $p_{\params}(\pitch \mid \wordsleft)$ and $p_{\params}(\pitch \mid \wordsbi)$, respectively.
During training, we fine-tune the combined model, not just the MDN network.
Further, when words are tokenized into multiple parts, we use the representation of the final token.
We refer to these methods as \mdnpast (for our mGPT-based estimates) and \mdnbi (for our mBERT-based estimates).
For a concurrent paper that uses similar methods to more fully investigate the relationship between context length and prosody, see \citet{regev2025time}.

As both mGPT and mBERT have access to context, these methods may use non-lexical properties of the context that affect pitch to make their predictions. For example, although English is not a tonal language,  punctuation (e.g., question marks) or wh-words at the beginning of a phrase can provide strong cues to phrase-final pitch. 
Therefore, there may be nonzero \mi between pitch and mBERT representations, even for non-tonal languages. 
We, thus, make a prediction: 
because these representations contain more than just lexical information, there should be less clear differences between tonal and non-tonal languages when mGPT and mBERT are used. 
This is in contrast to the fastText and KDE models, which we expect to bear out the predictions given in \Cref{hyp:ordering}.

\newcommand{\sa}{\textcolor{AccentColor}{SA}}
\newcommand{\pa}{\textcolor{PitchColor}{PA}}
\newcommand{\tonal}{\textcolor{WordsColor}{Tonal}}

\begin{table}
\setlength\extrarowheight{-4pt}
\setlength{\tabcolsep}{4pt}
\centering
\footnotesize
\resizebox{\columnwidth}{!}{%
\begin{tabular}{r|cccrrrrr}
\toprule
    Language & Tag & Type & Family & Hours&  Tokens & Types & Speakers \\
\midrule
    German & \texttt{DE} & \sa & Indo-Euro. & 8.6 & 47819 & 13519 & 338 \\
    English & \texttt{EN} & \sa & Indo-Euro.  & 7.8 & 47670 & 10930 & 557 \\
    French & \texttt{FR} & \sa & Indo-Euro. & 7.4 & 27974 & 8062 & 260 \\
    Italian & \texttt{IT}& \sa & Indo-Euro.  & 8.7 & 39413 & 10937 & 1641 \\
    Japanese & \texttt{JA} & \pa & Japonic & 6.4 & 54866 & 6434 & 896 \\
    Swedish & \texttt{SV} & \pa & Indo-Euro. & 6.6 & 38761 & 8002 & 461 \\
    Vietnamese & \texttt{VI} & \tonal & Austroasiatic & 5.9 & 37838 & 2468 & 130 \\
    Thai & \texttt{TH} & \tonal & Kra-Dai & 6.8 & 42153 & 4315 & 1749 \\
    Cantonese & \texttt{YUE} & \tonal & Sino Tibetan & 6.5 & 37380 & 6753 & 747 \\
    Mandarin & \texttt{ZH} & \tonal & Sino Tibetan & 7.9 & 36729 & 12547 & 1723 \\ 
\bottomrule
\end{tabular}
}
\caption{Overview of the languages and dataset used in this study. \sa = Stress Accent, \pa = Pitch Accent.}
\label{tab:languages-data}
\end{table}


\subsection{Dataset}

We use the Common Voice dataset \citep{ardila-etal-2020-common}, a multilingual corpus that contains paired text--audio samples from contributors reading individual sentences out loud.\footnote{The dataset is released under a Creative Commons Attribution Share-Alike license.} Samples are rated by other contributors who assign them either a thumbs-up or a thumbs-down. The validated portion of the dataset that we use includes only sentences whose first two ratings are up-votes. We select data from ten languages, across five different language families, representing a range of stress-accent, pitch-accent, and tonal languages (\cref{tab:languages-data}). We sample $5,000$ sentences per language for consistency, based on the language with the fewest validated sentences. In order to extract word-level prosodic features we align each sentence's audio to its text at the word level using the Montreal Forced Aligner \citep[MFA;][]{mcauliffe2017montreal}. For our Sino-Tibetan languages (Mandarin and Cantonese) we use two different tokenization or word-grouping schemes. In one both MFA alignment and NLP tokenization use characters as input units (this is tagged with \texttt{(chr)} in figures), and in the other MFA aligns audio to words, and NLP tools tokenize sentences into words using their default tokenizer.

The details of each language are given in \Cref{tab:languages-data}. Although this is a relatively modest sample of languages, it includes all languages in Common Voice that met the criteria for our data preparation---i.e., they have at least 5,000 sentences of validated data, an existing MFA model, and are well supported in the training data of our two neural LMs mBERT and mGPT (see \Cref{sec:mi_pitch_sentence}).

\subsection{Representation of Pitch} \label{sec:pitch-repr}

Representing the pitch curve of a word presents substantial challenges: We want to find a relatively low-dimensional representation space, but one that can still capture the complexities of pitch contours across languages, which may, for example, contain rising and falling elements on a single word. To do so, we use the preprocessing methods given in \citet{suni2017estimation} to extract the fundamental frequency, $f_0$ from the raw waveforms from each aligned word segment, and to remove outliers. We apply interpolation to create a smooth $f_0$ curve across moments where no pitch is being produced, for example, during unvoiced consonants. Once it has been extracted, we resample the $f_0$ curves to 100 points and parameterize them with the first four coefficients of a discrete cosine transform (DCT). The objective of our prosodic pipeline, therefore, is to estimate the four coefficients of the DCT pitch representation.

\begin{figure*}[t]
    \centering
    \includegraphics[width=\linewidth]{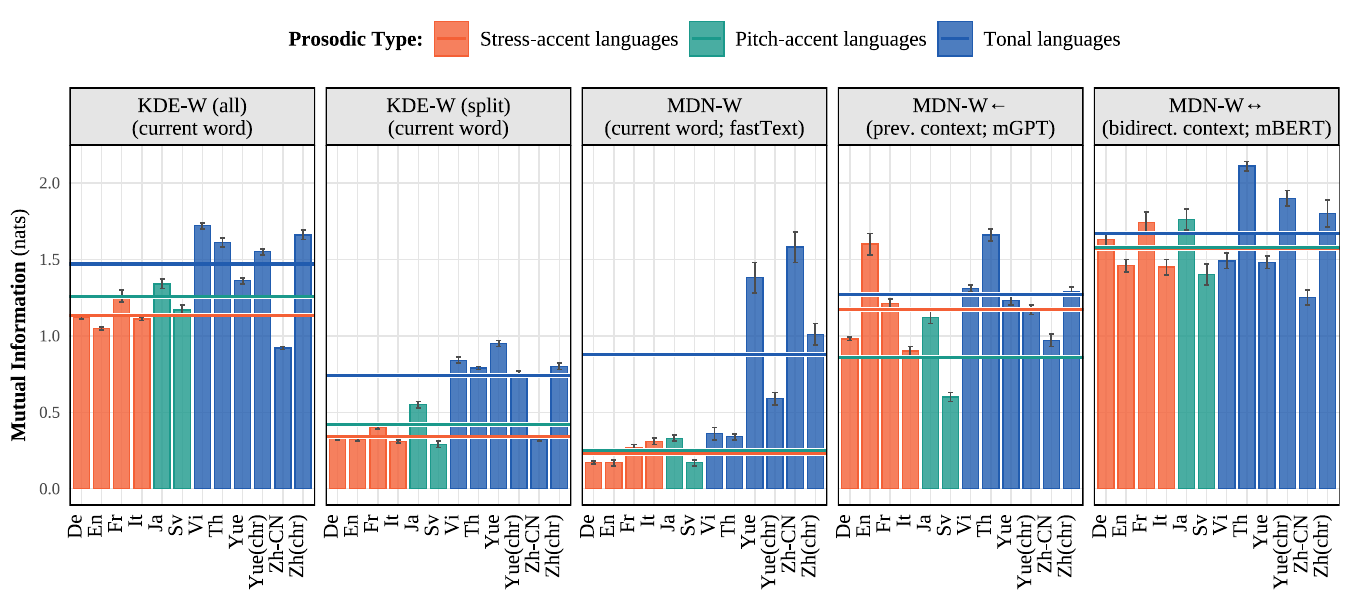}
    \caption{\textbf{Main Results:} Mutual information between pitch and text across languages. Lines show within typological group averages. Error bars show standard deviations from Monte Carlo resampling (\kdeall, \kdesplit) or 5-fold cross-validation (\mdnword, \mdnpast, \mdnbi). We find that tonal languages have higher \mi on average compared to stress-accent and pitch-accent languages.}
    \label{fig:main_results}
\end{figure*}

\section{Results} \label{sec:results}

\subsection{Main Results}

\paragraph{Mutual Information:} The results of our experiment are visualized in \Cref{fig:main_results}, with our different representations of text across the different facets. Horizontal bars show within typological group averages. The data support the typological ordering hypothesis: We observe higher \mi in tonal languages compared to non-tonal languages, for all of our estimation methods. Additionally, we find evidence supporting the tonal >> pitch-accent >> stress-accent hierarchy, especially for our \kdeall, \kdesplit, and \mdnword methods. The ordering is not present for \mdnpast, where stress-accent languages have higher average \mi than pitch-accent languages, or for \mdnbi, where stress- and pitch-accent languages have almost identical \mi.
To verify the visual trend in the results, we conducted a Jonckheere trend test \citep{jonckheere1954test}. This is a non-parametric method that tests whether samples are drawn from different populations with an \emph{a priori} ordering, compared to a null hypothesis where samples are all drawn from the same population. We use the implementation provided by the \texttt{clinfun} package in \texttt{R}, and approximate our $p$-values using $10,000$ permutations. Our test is significant for \kdeall ($p<0.05$), \kdesplit ($p<0.05$), and \mdnword ($p<0.01$) methods, but not for \mdnpast or \mdnbi, confirming the visual trend. 

Following the logic outline in \Cref{sec:mi_pitch_sentence}, we observe the greatest separation between tonal and non-tonal languages when using estimation techniques that do not take context into account (i.e., \mdnword and our two KDE-based methods). While estimation methods that incorporate longer context tend to have higher mutual information on average, these methods collapse the difference between typological groups. For example, using \mdnbi, we find the highest average \mi of any model, but we also find almost no difference between tonal and stress-accent languages, in terms of group averages. We suspect this is because \mdnbi, using BERT's bidirectional context, is capable of representing non-lexical information that can be useful for predicting pitch even in non-tonal languages, e.g., whether a given sentence is a question.

Interestingly, even though prosodic type behavior is consistent across models (i.e., tonal languages always have the highest \mi), within each prosodic type, models show variability. For example, our KDE-based methods both suggest that French is the stress-accent language with the highest \mi between pitch and lexical item. However, when using \mdnword, we find the highest \mi for Italian and for \mdnpast, English.
One possibility is that the different ways we represent context between these models lead to different amounts of \mi. 
We return to this point in the larger context of our \emph{gradient} vs. \emph{categorical} hypotheses in the discussion.

\begin{figure*}[t]
    \centering
    \includegraphics[width=\linewidth]{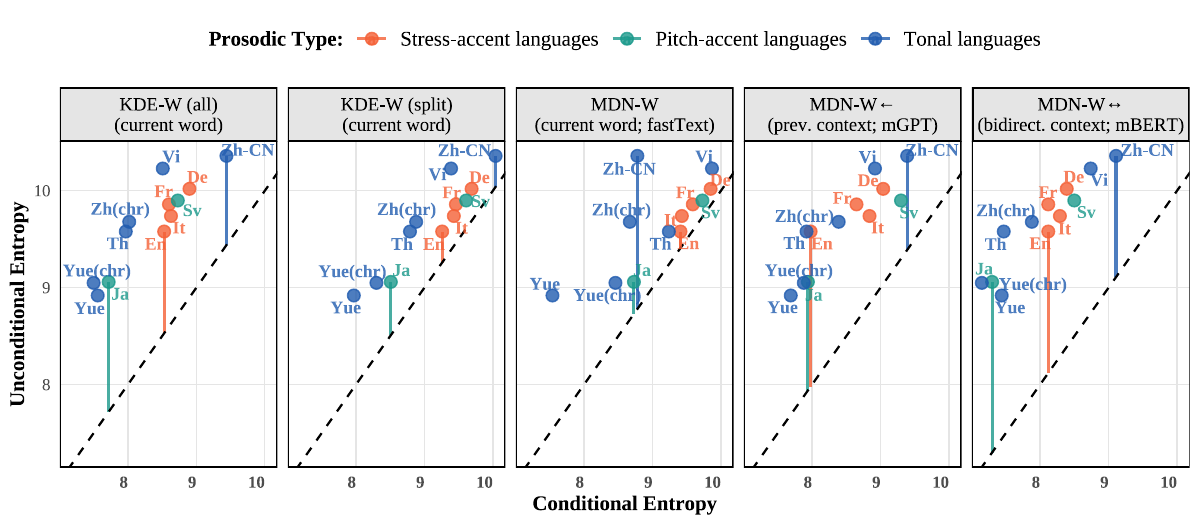}
    \caption{\textbf{Main Results Separated into Conditional/Unconditional Entropy:} Dashed line shows the $y=x$ line. Points show individual languages. Colored lines for En, Ja, and Zh visualize the \mi of these languages, which is the points' vertical distance from the dotted line.}
    \label{fig:entropy_results}
\end{figure*}

\paragraph{Conditional and Unconditional Entropy:} To zoom in on these data further, \Cref{fig:entropy_results} shows the same results broken down into conditional and unconditional entropy. The difference between these two is the \mi, shown in \Cref{fig:main_results} and visualized here as the vertical distance to the $x=y$ line, which is plotted for English, Japanese, and Mandarin. Overall, we observe a relatively narrow range for both unconditional entropy (ranging from $9$--$10.5$ nats) and conditional entropy (ranging from $7$--$10$ nats) across languages. These data support recent studies showing that information-theoretic properties of human language exist within a narrow bandwidth \citep{bentz2017entropy, wilcox2023testing, pimentel-etal-2020-phonotactic} \looseness=-1

When looking at entropy instead of mutual information, we observe more consistency at the language level. For all methods, Vietnamese, Chinese, and German have higher entropy (both conditional and unconditional), and Japanese, Cantonese, Thai, and English have lower entropy. The overall amount of entropy present in a language does not follow typological patterns or even the complexity of a language's tonal system. Cantonese, which is traditionally analyzed as having nine tones, always has lower entropy values than Mandarin, which is typically analyzed as having only four.

\subsection{The Role of Phonotactic Complexity and Syllable Structure}

One potential worry with the above results is that the prosodic types might co-vary with other features that could impact our \mi estimation, in particular phonotactic complexity and syllable structure.\footnote{We thank an anonymous reviewer for raising this issue.} To alleviate concerns about these potential confounds, we conducted the following checks: First, to examine phonotactic complexity, we used the measure proposed in \citet{pimentel-etal-2020-phonotactic}. We found that stress-accent languages possess slightly more complexity than pitch-accent languages ($\mu=3.4, \sigma=0.26$ vs. $\mu=3.0, \sigma=0.01$). However, \citeauthor{pimentel-etal-2020-phonotactic} did not report results for our tonal languages. As a second source of data, therefore, we used the World Atlas of Language Structures \citep[WALS;][]{dryer2013wals} features of ``consonant inventory size'' and ``vowel inventory size'' as a proxy for phonotactic complexity. We find that all of our languages for which data is recorded have ``average'' consonant inventory sizes, except for Japanese, which is ``moderately small.'' In addition, all of our languages have ``large'' vowel inventories, except for Japanese and Mandarin, which are ``average.'' We take this to mean there are not large differences in these complexity measures by prosodic type.

Turning to syllable structure, using the WALS ``syllable structure'' feature we find that all of our stress-accent languages have ``complex'' syllables; and all of our pitch-accent and tone languages have ``moderately complex'' syllables. Given that we observe the biggest differences in \mi between pitch-accent and tone languages, and only minimal differences between pitch- and stress-accent languages, we do not think that syllable complexity is therefore responsible for differences in \mi.

\begin{figure*}[t]
    \centering
     \includegraphics[width=0.99\linewidth]{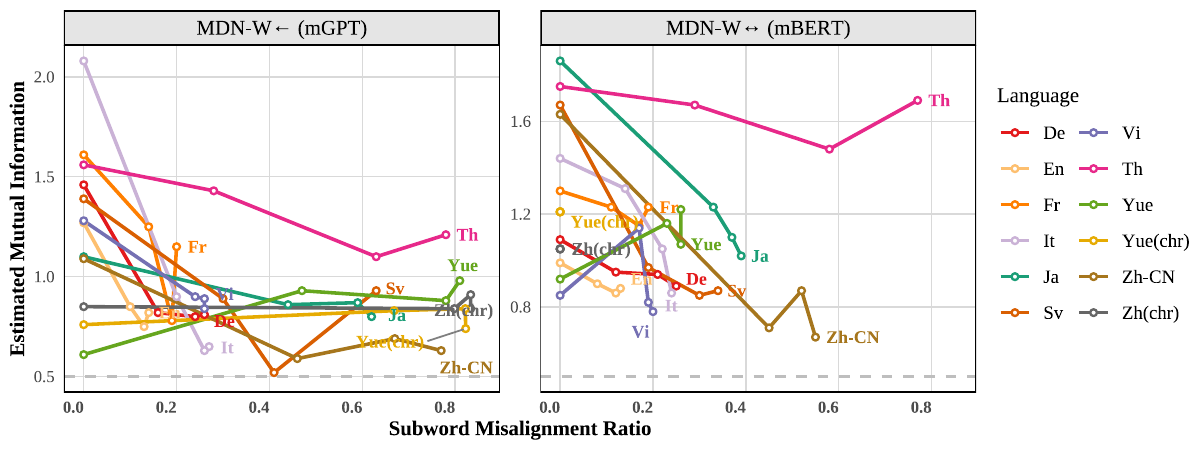}
    \caption{\textbf{Impact of tokenization on \mi estimation:} $x$-axis shows the proportion of words in our dataset tokenized into more than one token. Subsampling data to include only words with one token changes the estimated \mi.}
    \label{fig:subwords_tokenization}
\end{figure*}

\subsection{Effect of Subword Tokenization}

One difference between \kdeall, \kdesplit, and \mdnword, on one hand, and \mdnpast and \mdnbi, on the other hand, is that the LLMs that form the basis of the latter two methods (mBERT, mGPT) use subword tokenization schemes. For words that have multiple tokens, we used the embedding of the last token in the word during estimation. It's possible that this skews or biases our results.\footnote{In fact, \citet{lesci-etal-2025-causal} shows that an LM can assign the same word 2.7 times less probability if tokenised into two tokens compared to just one token.} Additionally, the number of single-token words varies across languages within our multilingual models, with English having more single-token words than the other languages. To investigate tokenization's impact, we took each of our initial datasets and subsetted them to include only words with $k$ or fewer tokens. We then re-ran our \mi estimation procedure using only the \mdnpast and \mdnbi methods. This resulted in datasets that were balanced in terms of tokens-per-word, but not in terms of total dataset size. \looseness=-1

The results are visualized in \Cref{fig:subwords_tokenization}. We see that as the percentage of multi-token words decreases, the \mi estimation changes, suggesting that, indeed, this impacts our results. However, the overall picture of the results remains the same---there is no clear separation between tonal, pitch-accent, and stress-accent languages using these models. Interestingly, as the tokens-per-word ratio decreases, the \mi increases for most (although not all) languages, suggesting that the \mi estimates in \Cref{fig:main_results} are slight underestimates. 
For additional presentation of these data see \Cref{app:tokenization}.

\section{Discussion}

Our experiments supported the typological ordering hypothesis, namely that tonal languages have higher \mi between pitch and text, followed by pitch-accent and stress-accent languages. The ordering of languages according to this prediction is relatively clean, especially for the tonal vs.\ non-tonal distinction. Among the KDE-based estimates, where we expect the separation to be the strongest, we found only one tonal language (Cantonese, word level) with a lower \mi than any stress-accent language. And with \mdnword, we found that all tonal languages had higher \mi than all stress-accent. Finally, we generally found that pitch-accent languages fell between tonal and stress-accent languages, as expected.

What do our results say about the status of categorical vs.\ gradient typological theories? On one hand, they could be construed to support the categorical prediction. Using our \mdnword method, we find a single amount of mutual information ($0.34$ nats) that separates all tonal from non-tonal languages. At the same time, our results demonstrate interesting gradient differences both between and within prosodic types. Firstly, it's not the case that languages are clearly separated into different modes based on typological type. For example, using our \mdnword method, there is far more variation in \mi \emph{within} tonal languages (ranging from $0.36$--$1.58$ nats) than \emph{between} tonal vs.\ stress-accent groups ($0.23$ vs.\ $0.88$ nats). Based on these considerations, we conclude that our data are more closely aligned with the gradient prediction, outlined in \Cref{sec:predictions}.

We close our theoretical discussion by clarifying the relationship between our definition of a $\pitch$-language and  \citeposs{greenberg2005universals} notion of an implicational universal. While implicational universals result in mutual information between linguistic properties, it is not possible to reduce such universals to \mi alone. To take one example, a well-studied implicational universal holds that VSO languages always have prepositions (as opposed to \emph{post}positions). This implies that there is mutual information between a language's word order and its adposition placement. However, if the implication was reversed---VSO implies postpositions---the amount of \mi would remain unchanged. Importantly, implicational universals specify \emph{how} features of a language covary, not just that they \emph{do} covary. Zooming out, we can say that implicational universals and $\pitch$-languages are a larger class of linguistic variation that implies \mi between linguistic features. Further characterizing how mutual information relates to known typological features is an important direction for future research.

Finally, a methodological point: This paper has focused on pitch; however, prosodic typologies operate across a broad range of dimensions. We want to stress that our methods are not only suitable for studying pitch: We initially framed our technical presentation in terms of abstract prosodic categories \pitch. While \pitch could in theory be all of prosody, it can also be just a single prosodic feature. One could equally well use our methods to examine length-based lexical distinctions, for example, in languages like Turkish. We hope others will build on the technical contributions offered here to study a broader range of prosodic phenomena.


\section*{Limitations}

One limitation of this work has to do with our dataset: First, the dataset is relatively small, with just $5,000$ sentences per language. Second, we did not control for the number of unique speakers in the dataset, meaning that some languages are overly represented by a single or handful of individuals. For example, our Thai data includes samples from $1,749$ speakers, whereas our Vietnamese data includes samples from just $130$ speakers. One other shortcoming of our dataset is that while our pitch-accent and tonal languages include data from multiple language families, our stress-accent data comes entirely from Indo-European languages. Finally, our dataset did not control for content, meaning the distribution of concepts and, therefore, words could vary substantially across languages. While collecting high-quality audio-text-aligned data across multiple languages is a difficult undertaking, assembling such a dataset and running similar analyses would be an excellent way to further validate the conclusions of this study.

\section*{Ethics Statement}

We foresee no obvious ethical problems with this research. Furthermore, we do not foresee any obvious risks with this research.

\bibliography{custom}

\appendix

\section{Subword Tokenization Follow-up Analysis}
\label{app:tokenization}

\begin{figure*}[t]
    \centering
    \begin{minipage}[c]{\textwidth}
    \includegraphics[width=0.9\linewidth]{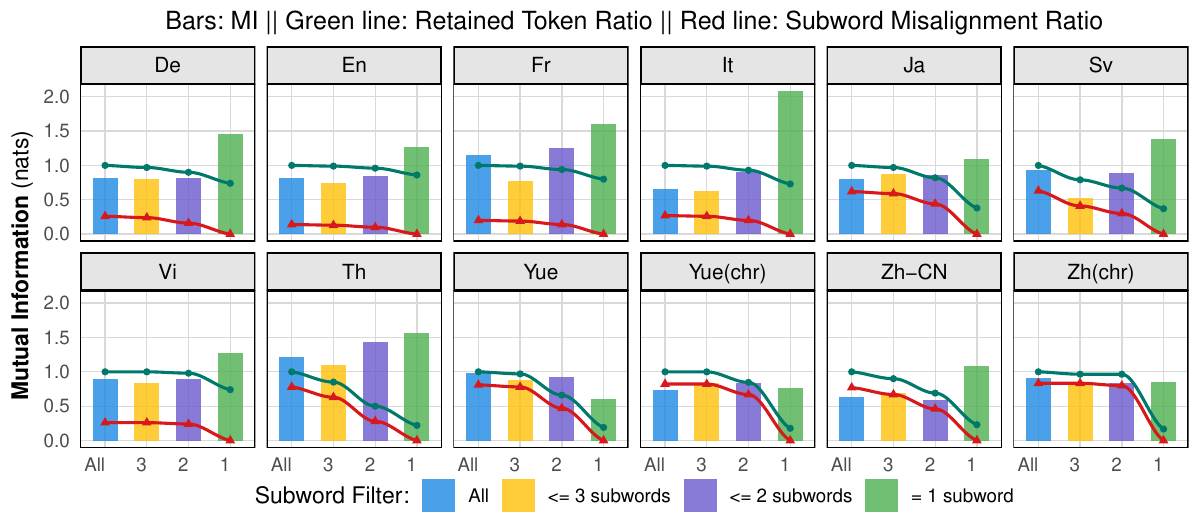}
    \subcaption{\mdnpast (mGPT)} \label{fig:mGPT}
    \end{minipage}
    \begin{minipage}[c]{\textwidth}
     \includegraphics[width=0.9\linewidth]{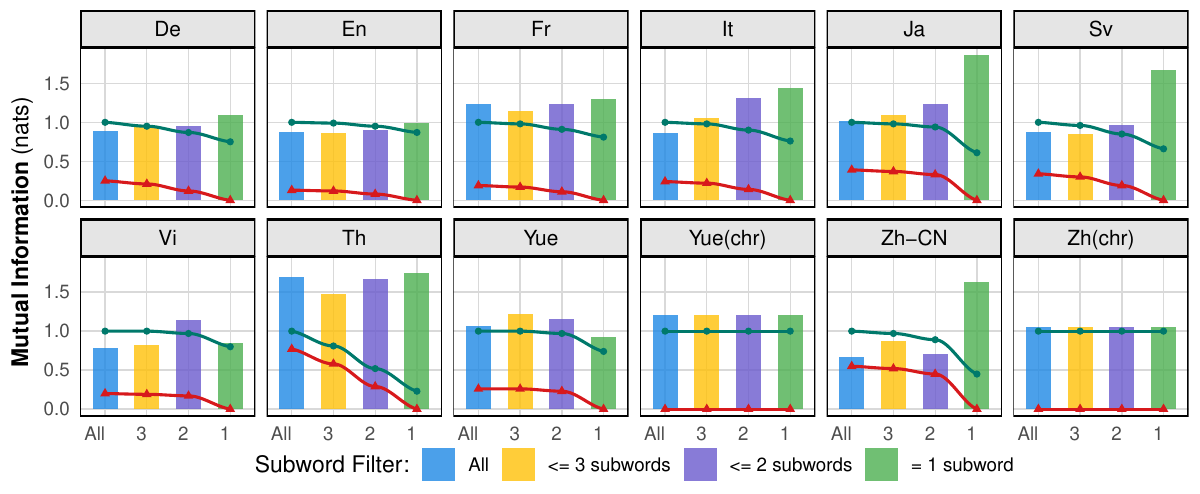}
     \subcaption{\mdnbi (mBERT)} \label{fig:mBERT}
     \end{minipage}
    \caption{\textbf{Fine-Grained Analysis of Subword Tokenization Effects on \mi Estimation:} 
    The x-axis represents subword filtering levels: ``All'' (no filtering), ``3'' (subsetted words with at most 3 subword tokens), ``2'' (at most 2 tokens), and ``1'' (only single-token words). Bars show estimated \mi, the green line represents the retained token ratio after subsetting, and the red line represents the misalignment ratio in the retained data.}
\label{fig:subwords_tokenization_fine_grained}
\end{figure*}

In this appendix, we present more fine-grained data concerning the impact of subword tokenization on our \mi estimation. These data are presented in \Cref{fig:subwords_tokenization_fine_grained}. We find that, in general, filtering out multi-token words increases \mi, implying that subword tokenization misalignment adds noise to the estimation procedure. In particular, \mi tends to be highest for our subsampled datasets that include words with only one token---the green bars to the left of each facet.  Cantonese (Yue) is an exception, for both of our models, likely due to its many single-character words.

Retained tokens (green line) and misalignment (red line) decrease as we subsample data. However, some languages like English, French, and German retain more data, while Chinese, Thai, and Swedish lose more, resulting in cleaner but smaller datasets for \mi estimation.

Languages also vary in initial misalignment (red lines). English has the lowest initial misalignment, while Chinese and Thai have more, leading to larger \mi gains after filtering and suggesting that \mi is likely underestimated in our main results for these languages when using our \mdnpast and \mdnbi techniques.

\section{Hyperparameter and Hyperparameter search}
\label{app:hyperparameter}
We performed a hyperparameter search using 5-fold cross-validation to tune the \mdnword model. The search space included:

\begin{itemize} \setlength{\itemsep}{0pt} 
\item Learning rate: 0.01, 0.001 
\item Dropout: 0.2, 0.5 
\item Hidden layers: 15, 20, 30 
\item Hidden units: 512, 1024
\end{itemize}

Models were trained for a maximum of 50 epochs using the AdamW optimizer with weight decay (L2 regularization = $0.001$) and early stopping (patience = 3) based on validation loss. The best hyperparameters were selected based on average performance across the 5 folds and evaluated on the test set.

For our \mdnpast, using mGPT (\texttt{ai-forever/mGPT}) and \mdnbi using mBERT (\texttt{bert-base-multilingual-cased}) models, we fine-tuned using AdamW (weight decay = 0.1), a learning rate of \(5.0 \times 10^{-5}\) with ReduceLROnPlateau (factor = 0.1, patience = 2), batch size 16 (effective 64), gradient clipping at 1.0, dropout of 0.1 (applied to the MLP head), and early stopping (patience = 3). For \mdnpast using mGPT, we fine-tune only the last eight transformer layers, freezing the rest for efficiency, resulting in 612M trainable parameters (out of 1.4B total). For \mdnbi using mBERT, all layers are fine-tuned, for 177M trainable parameters.

\end{document}